\documentclass[runningheads]{llncs}

\usepackage{graphicx}

\usepackage{graphicx}
\usepackage{subcaption}
\usepackage{hyperref}       
\usepackage{url}            
\usepackage{booktabs}       
\usepackage{amsfonts}       
\usepackage{nicefrac}       
\usepackage{microtype}      
\usepackage{color}
\usepackage{float}
\usepackage{authblk}
\usepackage{amsmath}
\usepackage{hyperref} 
\usepackage{fullpage}
\usepackage{lipsum}
\usepackage{float}
\usepackage[dvipsnames]{xcolor}
\usepackage{bm}
\usepackage{bbold}
\usepackage{amssymb}
\usepackage{longtable}
\usepackage{comment}
\usepackage{ragged2e}
\usepackage{array}
\usepackage{float}
\usepackage{tabularx}
\usepackage[normalem]{ulem}

\usepackage[numbers]{natbib}  
\makeatletter
\renewcommand{\@biblabel}[1]{[#1]}  

\makeatother
\usepackage{xr}
\renewcommand{\thetable}{\arabic{table}}

\usepackage{caption}
\newcolumntype{P}[1]{>{\centering\arraybackslash}p{#1}}
\newcolumntype{B}[1]{>{\justifying\parindent=0pt\arraybackslash}p{#1}}

\begin{document}

\title{Medical Imaging AI Competitions Lack Fairness}

\titlerunning{Medical Imaging AI Competitions Lack Fairness}

\author{Annika Reinke\inst{1,2}\footnote{Corresponding author: \email{a.reinke@dkfz-heidelberg.de}}, Evangelia Christodoulou\inst{1}, Sthuthi Sadananda\inst{3}, A. Emre Kavur\inst{1}, Khrystyna Faryna\inst{4, 5}, Daan Schouten\inst{4, 5}, Bennett A. Landman\inst{6, 7}, Carole Sudre\inst{8, 9, 10}, Olivier Colliot\inst{11}, Nicholas Heller\inst{12}, Sophie Loizillon\inst{11}, Martin Maška\inst{13}, Maëlys Solal\inst{11}, Arya Yazdan-Panah\inst{11}, Vilma Bozgo\inst{4}, Ömer Sümer\inst{14}, Siem de Jong\inst{4}, Sophie Fischer\inst{15}, Michal Kozubek\inst{13}, Tim Rädsch\inst{1, 2, 16, 17}, Nadim Hammoud\inst{1}, Fruzsina Molnár-Gábor\inst{18}, Steven Hicks\inst{19}, Michael A. Riegler\inst{20}, Anindo Saha\inst{21}, Vajira Thambawita\inst{19}, Pal Halvorsen\inst{19}, Amelia Jiménez-Sánchez\inst{22, 23}, Qingyang Yang\inst{1}, Veronika Cheplygina\inst{23}, Sabrina Bottazzi\inst{24}, Alexander Seitel\inst{1, 33}, Spyridon Bakas\inst{25, 26, 27, 28, 29, 30}, Alexandros Karargyris\inst{31}, Kiran Vaidhya Venkadesh\inst{4}, Bram van Ginneken\inst{4, 32}, Lena Maier-Hein\inst{1, 2, 33, 34, 35, 36}}

\institute{
German Cancer Research Center (DKFZ) Heidelberg, Division of Intelligent Medical Systems, Germany  
 \and DKFZ Heidelberg, Helmholtz Imaging, Germany
 \and Digital Oncology Program, National Center for Tumor Diseases (NCT) Heidelberg, Germany
 \and Department of Pathology, Radboud University Medical Center, Nijmegen, The Netherlands
 \and Oncode Institute, Utrecht, The Netherlands
 \and Vanderbilt University, Nashville, Tennessee, USA
 \and Vanderbilt University Medical Center, Nashville, Tennessee, USA
 \and MRC Unit for Lifelong Health and Ageing at UCL, Institute of Cardiovascular Science, University College London, London, UK
 \and Dementia Research Centre, UCL Queen Square Institute of Neurology, University College London, London, UK
 \and School of Biomedical Engineering, King’s College, London, UK
 \and Sorbonne Université, Institut du Cerveau -- Paris Brain Institute -- ICM, CNRS, Inria, Inserm, AP-HP, Hôpital de la Pitié-Salpêtrière, Paris, France
 \and Department of Urology, Cleveland Clinic, Cleveland, USA
 \and Centre for Biomedical Image Analysis, Faculty of Informatics, Masaryk University, Brno, Czech Republic
 \and DKFZ Heidelberg, Division of Computational Genomics and Systems Genetics, Germany
 \and Department of Computer Science, University of Oxford, Oxford, United Kingdom
 \and Engineering Faculty, Heidelberg University, Germany
 \and School of Computation, Information and Technology, TUM, Germany
 \and Faculty of Law / BioQuant, Heidelberg University, Heidelberg, Germany
 \and The Department of Holistic Systems (HOST), SimulaMet,Oslo, Norway
 \and AI Safety Department, SimulaMet, Oslo, Norway
 \and Department of Medical Imaging, Radboud University Medical Center, Nijmegen, the Netherlands
 \and Artificial Intelligence in Medicine Lab (BCN-AIM), Departament de Matemàtiques i Informàtica, Universitat de Barcelona, Barcelona, Spain
 \and Section of Data, Systems, and Robotics, IT University of Copenhagen (ITU), Copenhagen, Denmark
 \and Escuela de Ciencia y Tecnología, Universidad Nacional de San Martín, San Martín, Buenos Aires, Argentina
 \and Division of Computational Pathology, Department of Pathology \& Laboratory Medicine, Indiana University School of Medicine, Indianapolis, USA
 \and Indiana University Melvin and Bren Simon Comprehensive Cancer Center, Indianapolis, USA
 \and Department of Radiology \& Imaging Sciences, Indiana University School of Medicine, Indianapolis, USA
 \and Department of Biostatistics \& Health Data Science, Indiana University School of Medicine, Indianapolis, USA
 \and Department of Neurological Surgery, Indiana University School of Medicine, Indianapolis, USA
 \and Department of Computer Science, Luddy School of Informatics, Computing, and Engineering, Indiana University, Indianapolis, USA
 \and Medical working group, MLCommons, Dover, USA
 \and Plain Medical, Nijmegen, The Netherlands
 \and National Center for Tumor Diseases (NCT), NCT Heidelberg, a partnership between DKFZ and University Hospital Heidelberg, Heidelberg, Germany
 \and Heidelberg University Hospital, Surgical Clinic, Surgical AI Research Group, Heidelberg, Germany
 \and Faculty of Mathematics and Computer Sciences, Heidelberg University, Heidelberg, Germany
 \and Mohamed Bin Zayed University of Artificial Intelligence (MBZUAI), Abu Dhabi, UAE
 }

  \authorrunning{A. Reinke et al.}

\maketitle

\section{Abstract}
Benchmarking competitions are central to the development of artificial intelligence (AI) in medical imaging, defining performance standards and shaping methodological progress. However, it remains unclear whether these benchmarks provide data that are sufficiently representative, accessible, and reusable to support clinically meaningful AI. We assess fairness along two complementary dimensions: (1) whether challenge datasets capture the diversity of real-world clinical data, and (2) whether they are accessible and legally reusable in line with the FAIR principles. We systematically studied 249 biomedical image analysis challenges comprising 458 tasks across 19 imaging modalities. Here, we show limited representation with respect to geographic location, imaging modalities, and problem types, raising concerns about whether current benchmarks reflect real-world clinical diversity. Challenge datasets were frequently constrained by restrictive or ambiguous access conditions, inconsistent or non-compliant licensing practices, and incomplete documentation, limiting reproducibility and long-term reuse. These shortcomings expose foundational fairness limitations in our benchmarking ecosystem and highlight a disconnect between leaderboard success and clinical relevance.  
\setcounter{footnote}{0}

\section{Introduction}
Artificial intelligence (AI) has shown remarkable potential in medicine, such as disease detection, diagnosis, and treatment planning across imaging modalities. AI promises to transform medical imaging, yet few algorithms have reached clinical use \cite{panch2019inconvenient, christodoulou2025false}. This gap reflects a persistent lack of rigorous, real-world validation and benchmarking. In AI-based healthcare, benchmarking has become central to algorithm evaluation, enabling standardized comparisons and guiding methodological development. Over the past two decades, international biomedical image analysis competitions, known as challenges, have become central to AI benchmarking in medicine \cite{maier2018rankings, antonelli2022medical}. Challenges enable fair comparisons by evaluating algorithms under identical, controlled conditions, producing rankings that define the state of the art for a given research question. With landmark contributions such as nnU-net \cite{isensee2021nnu} emerging from these competitions, challenges have shaped community standards, publication practices, and perceptions of state-of-the-art performance. In some cases, they have led to clinical applications, such as MammoScreen, which originated from the Digital Mammography DREAM Challenge and was subsequently developed into an FDA-cleared clinical product\footnote{\url{https://www.accessdata.fda.gov/scripts/cdrh/cfdocs/cfpmn/pmn.cfm?ID=K192854}}. While many factors contribute to the quality and impact of challenges \cite{maier2018rankings}, a central aspect is their data.

Challenge datasets not only form the basis for ranking algorithms, they are widely reused as reference benchmarks in subsequent research. As such, they have become integral to scientific progress. As data drive research \cite{varoquaux2022machine}, future researchers will train, tune, and validate their methods on these datasets. Therefore, the quality of challenge data and its documentation is crucial for ensuring that benchmarking outcomes are both scientifically meaningful and clinically translatable. Because these datasets are widely reused for method development and evaluation, understanding their characteristics is a prerequisite for studying how benchmarking datasets may influence future methodological research.

Benchmarking requires data that are not only available to the research community but also representative and reusable. These aspects reflect two core dimensions of fairness in biomedical AI: 
\begin{enumerate}
    \item Fairness in representativeness concerns whether datasets reflect real-world conditions, such as geographic or clinical variation.
    \item Fairness in reuse relates to whether challenge data can be accessed and reused, including accessibility conditions, licensing clarity, and documentation quality. These aspects align with the FAIR (Findability, Accessibility, Interoperability, and Reusability) data principles \cite{wilkinson2016fair}\footnote{In this work, the FAIR principles are used as a conceptual reference, with a specific focus on Accessibility and Reusability. FAIR is not interpreted as a legal or normative standard; instead, we empirically assess practical barriers affecting access to and reuse of challenge data.}, with our analysis focusing on Accessibility and Reusability. Accessibility refers to whether data can be obtained without unnecessary barriers or opaque approval processes, and Reusability addresses whether data can be legally and practically reused, redistributed, and understood based on their documentation.
\end{enumerate}

Previous work has investigated biomedical image analysis challenges, including challenge design, ranking robustness, and reproducibility \cite{maier2018rankings, reinke2018exploit, reinke2023challenge}. Concerns regarding the representativeness and quality of challenge datasets have been raised within the biomedical image analysis community, for example in international community surveys \cite{maier2018rankings}. However, the prevalence of these data-related issues has not yet been systematically assessed across biomedical image analysis challenges. Building on our previous work on trustworthy validation in medical imaging (e.g., \cite{maier2018rankings, maier2024metrics}), this study -- initiated by the Medical Image Computing and Computer Assisted Interventions (MICCAI) Special Interest Group for Challenges, the Medical Open Network for AI (MONAI) Benchmarking and Evaluation Working Group, Helmholtz Imaging, and the Grand Challenge team -- provides the first quantitative assessment of how fairly and transparently biomedical AI challenges share and document their data. 

\begin{figure}[H]
    \centering
    \includegraphics[width=1\linewidth]{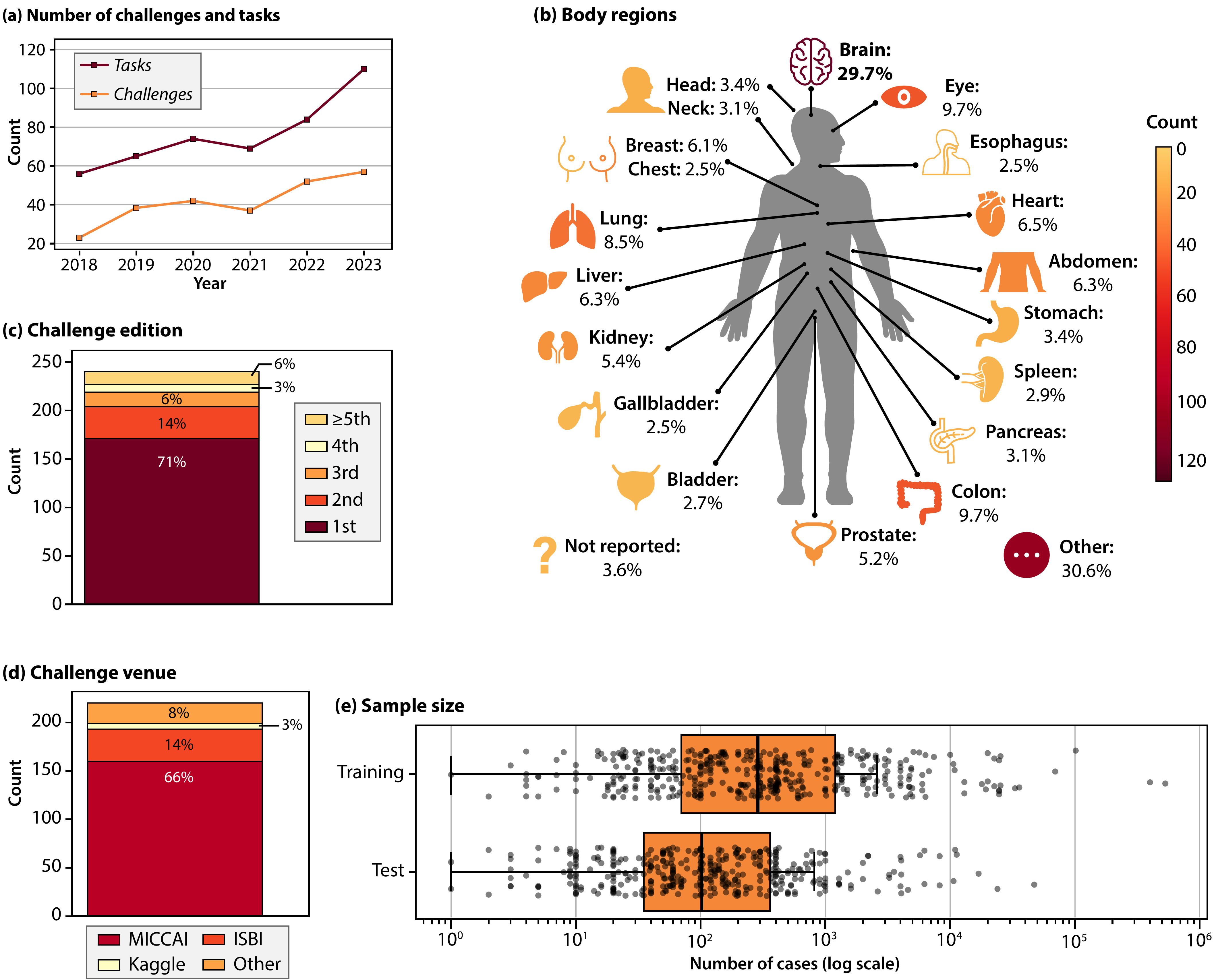}
    \caption{\textbf{Overview of biomedical image analysis challenges.} (a) Number of challenges and tasks per year. (b) Body region. (c) Challenge edition, i.e., was the challenge newly introduced or an iteration of previous versions. (d) Challenge venue or conference. (e) Sample size of training and test datasets. Each data point represents one challenge task with a reported dataset size (training: n = 429 tasks; test: n = 394 tasks). Box plots indicate the median (center line), 25th and 75th percentiles (box bounds), and the most extreme values within 1.5 × the interquartile range (whiskers).}
    \label{fig:fig1}
\end{figure}

The goal of this work is not to define how biomedical image analysis challenges should be organized. Instead, we aim to raise awareness of systematic limitations in challenge datasets, provide an empirical basis for discussing these limitations, and encourage improvements where feasible. We make the following key contributions to understanding fairness in biomedical AI benchmarking:
\begin{enumerate}
    \item \textbf{Large-scale analysis of biomedical image analysis challenges:} We conducted a large-scale systematic review of recent biomedical imaging challenges, spanning 458 medical imaging AI tasks, across 19 modalities (Fig.~\ref{fig:fig1});
    \item \textbf{Empirical evidence of limited representation in challenge datasets:} Our analysis reveals substantial overrepresentation of certain geographic regions and imaging modalities, alongside a predominance of specific problem types and anatomical regions (Figs.~\ref{fig:fig1}b, \ref{fig:fig2}--\ref{fig:fig3});
    \item \textbf{Critical flaws in data accessibility, licensing, and documentation:} We expose  widespread issues in data accessibility, licensing practices, and documentation quality, including inconsistent, unclear, or non-compliant terms and restrictive access conditions (Fig.~\ref{fig:fig4}, Tab.~\ref{tab:inconsistencies}).
\end{enumerate}

\section{Results}
Benchmarking competitions have become central to AI development in medicine, shaping model design, benchmarking practices, and identifying potential for translation into real-world clinical practice. To assess whether medical imaging AI challenges fulfill this role, we conducted a large-scale systematic analysis based on double-blind annotations from 30 observers. Our findings reveal substantial differences in the representation across key dataset characteristics, restrictions in data accessibility and reusability, as well as documentation gaps that limit the fairness, transparency, and reliability of challenge data and the research that builds upon them.

\newpage

\subsection{Current landscape of biomedical image analysis challenges}
We identified 249 challenges comprising 458 tasks between 2018 and 2023 (Fig.~\ref{fig:fig1}). On average, challenges had 1.7 tasks (median: 1, IQR: 1, maximum: 10). Most challenges were newly introduced (71\%), whereas only 6\% had reached their fifth or higher edition. The majority of challenges were hosted at the International Conference on Medical Image Computing and Computer Assisted Intervention (MICCAI; 66\%), followed by the IEEE International Symposium on Biomedical Imaging (ISBI; 14\%). Median dataset sizes differed across venues. While challenges hosted at MICCAI (276/100 training/test cases) and ISBI (400/113) were typically based on datasets of similar scale, challenges at NeurIPS (2320/1311) and RSNA (1251/570) generally used substantially larger datasets. Across all entries, information was primarily obtained from  challenge websites (41\%), challenge papers (24\%), and registered challenge design documents (23\%).

The number of participating teams varied considerably (median: 12; maximum: 3,308). Participation also differed substantially between stages, with a median ratio of participating teams between the test and final phases 2.3:1. The number of participating teams varied considerably across venues. Kaggle challenges had by far the highest average and median numbers of participants (mean: 1,289; median: 1,347). In contrast, challenges at venues such as MICCAI, the Conference on Neural Information Processing Systems (NeurIPS), ISBI, and Medical Imaging with Deep Learning (MIDL) tended to involve fewer teams (e.g., median $<$ 20; Suppl. Tab.~\ref{tab:suppl-tab1}).

Most challenges required either algorithm or code submission (49\%) or result submission (46\%). Historically, result submission massively dominated until 2021, accounting for more than 70\% of tasks yearly. Since then, Since then, algorithm (or code) submission has become the prevailing requirement, with 75\% of tasks in 2023.

Dataset sizes varied considerably (Fig.~\ref{fig:fig1}e and Suppl. Tab.~\ref{tab:suppl-tab4}; training/validation/test minimum: 0/1/1; median: 288/80/102; maximum: 530,706/18,368/47,227). The median ratio of training and test cases was 2.2:1. The median number of annotators was 3 (maximum: 92), while the median number of annotators per case was 2 (maximum: 7). Experts were involved in the annotation process in 47\% of tasks. Annotators with intermediate experience were involved in 26\% of tasks, and trainees, including students, in 19\%. Expert supervision or final review was reported for 31\% of tasks. Only 1\% of tasks used a professional annotation company. Annotator expertise was unclear or not reported for 27\% of tasks.

The most common awards were monetary prizes (39\% of tasks). Among those with monetary awards, the exact prize amount ranged from 117 EUR to 120,000 EUR (mean: 6,046 EUR, median: 1,814 EUR, IQR: 3,380 EUR). For 31\% of challenges, submission after the official challenge ended remained possible. Overall, 66\% of challenges had an associated publication (57\% challenge paper, 25\% data publication, 16\% both). The median citations per year was 20 (maximum: 444) for challenge papers and 22 for data papers (maximum: 141). The median publication delay between challenge opening and challenge paper publication was 1.6 years (maximum: 6.6 years).

\subsection{Challenge datasets show limited geographic and thematic representation}
The ability of challenges to support generalizable AI development depends on the diversity and representativeness of modalities, tasks and patient characteristics. However, our findings reveal pronounced differences in the representation of geographic regions, problem categories, imaging modalities, and anatomical regions. 

\begin{figure}[H]
    \centering
    \includegraphics[width=1\linewidth]{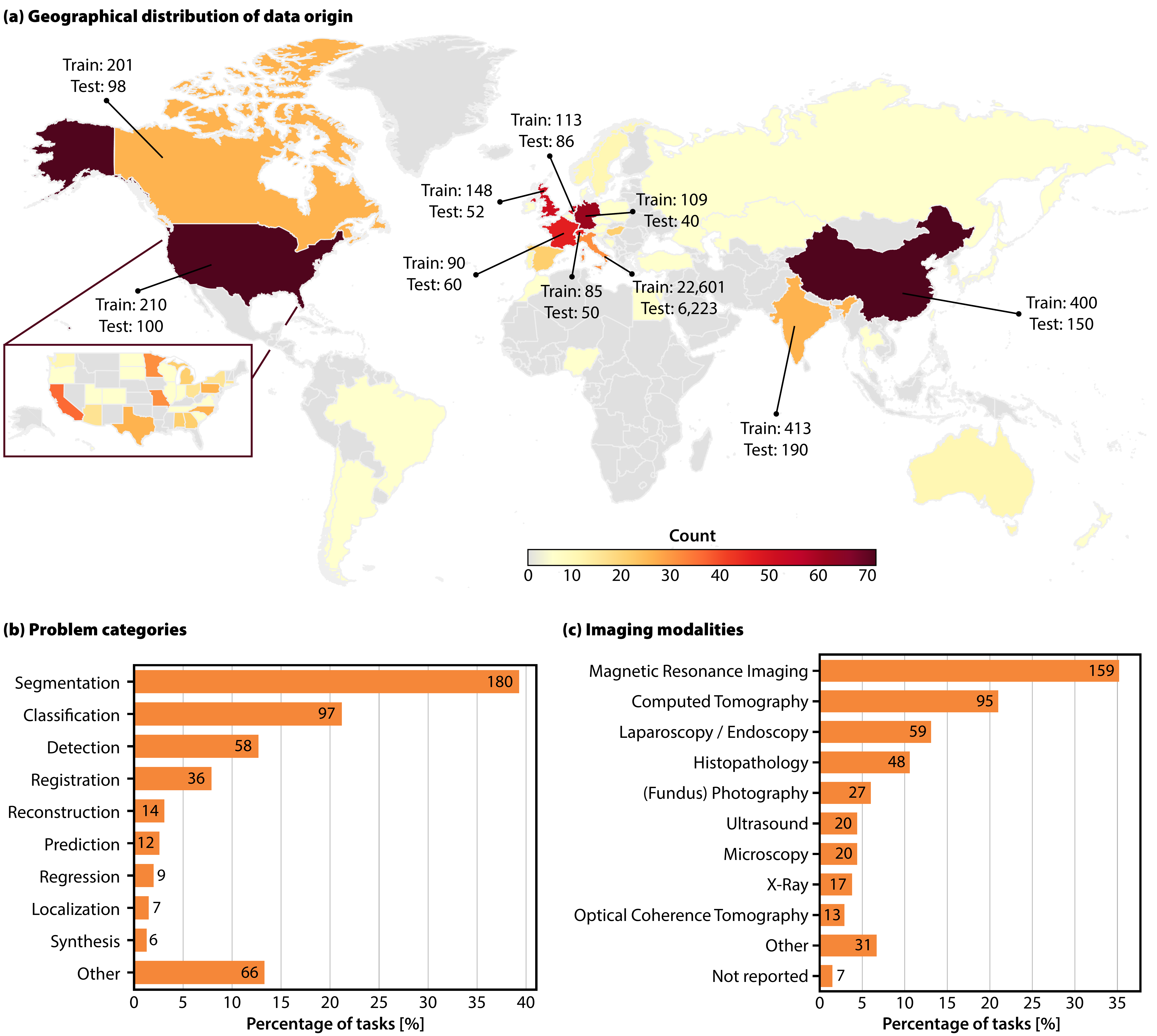}
    \caption{\textbf{Medical imaging AI challenges are are dominated by specific} (a) geographical origin (Northern America, China, Europe), (b) problem categories (segmentation), and (c) imaging modalities (Magnetic Resonance Imaging; MRI). For the most frequently represented countries, (a) additionally reports the median training and test set sizes.}
    \label{fig:fig2}
\end{figure}

\begin{figure}[H]
    \centering
    \includegraphics[width=\linewidth]{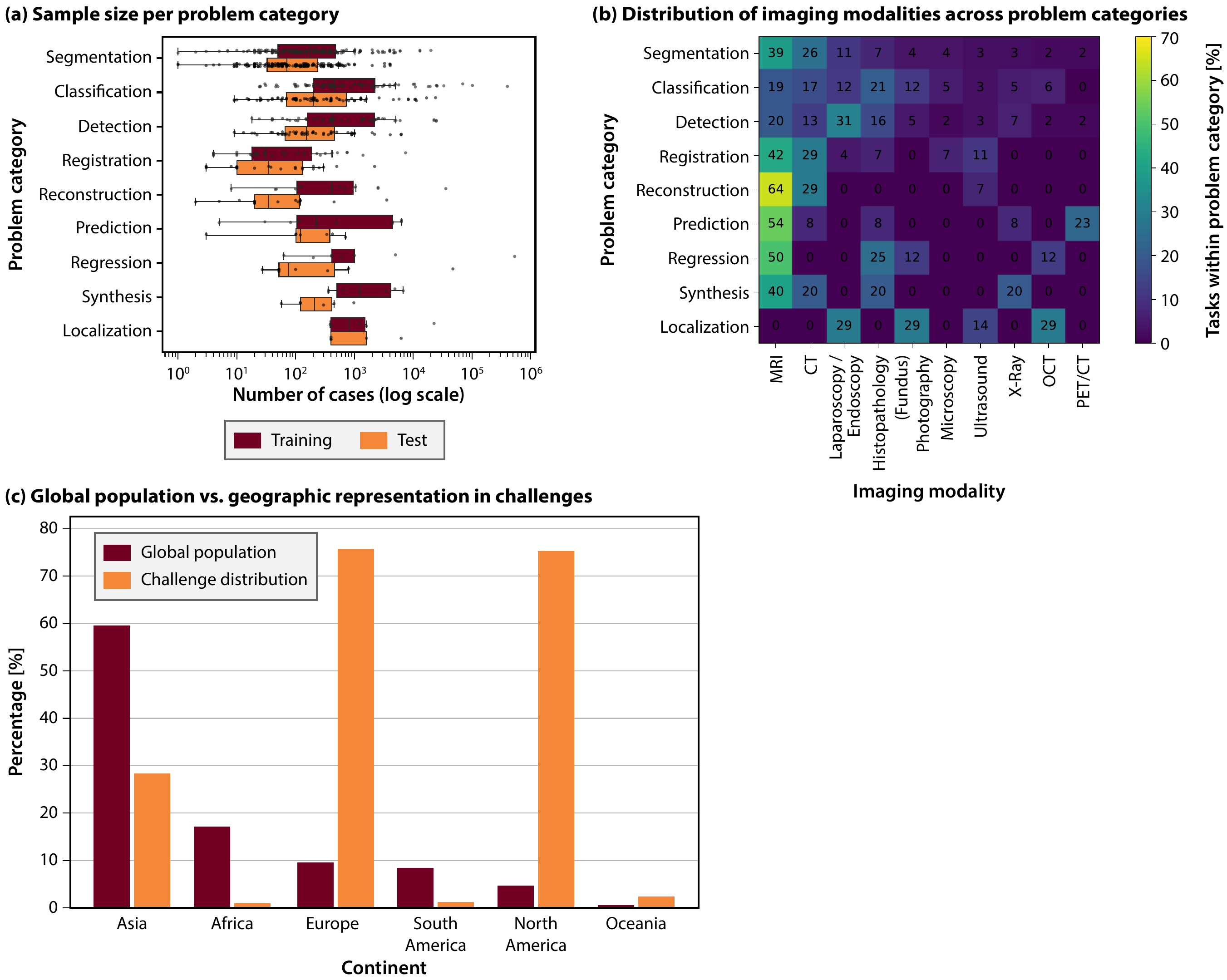}
    \caption{\textbf{Challenge distributions are associated with dataset characteristics and show substantial deviations from external reference distributions.} (a) Training and test set sizes across problem categories. Boxes represent the interquartile range with the median indicated by the center line; whiskers extend to 1.5 times the interquartile range. Individual values per task are shown as jittered dots. Sample sizes (training/test): segmentation (n = 173/166 tasks), classification (n = 86/81), detection (n = 53/52), registration (n = 35/25), reconstruction (n = 14/14), prediction (n = 12/7), regression (n = 9/8), synthesis (n = 7/6), and localization (n = 6/5). (b) Distribution of imaging modalities across problem categories. Values indicate the percentage of tasks within each problem category assigned to the respective imaging modality. Percentages are normalized per problem category. (c) Geographic representation of challenge data compared with the global population distribution (numbers based on \cite{sadigov2022rapid}). Values indicate the percentage of tasks containing data from the respective continent and the corresponding share of the global population. As individual tasks may include data from multiple continents, challenge percentages may exceed 100\%.}
    \label{fig:fig3}
\end{figure}

70\% of tasks used data originated from the United States, followed by China (16\%), Germany (13\%), the Netherlands (10\%), and the United Kingdom (10\%) (Fig.~\ref{fig:fig2}a), while regions such as Oceania (2.4\%), South America (1.3\%), or Africa (1.0\%) were rarely represented. Relative to the global population distribution, North America and Europe were overrepresented by factors of 15.9 and 7.9, whereas Africa (0.1), South America (0.2), and Asia (0.5) were underrepresented relative to the global population (Fig.~\ref{fig:fig3}c, Suppl. Tab.~\ref{tab:suppl-tab2}). Median dataset size varied across countries, ranging from 17/10 training/test cases in Japan to 22,601/6,223 in Italy (Fig.~\ref{fig:fig2}a; Suppl. Tab.~\ref{tab:suppl-tab5}).

The most frequently represented problem categories were segmentation (39\% of tasks), classification (21\%), and detection (13\%), while other problem types accounted for smaller proportions (Fig.~\ref{fig:fig2}b). Median dataset sizes varied across problem categories, ranging from 30/35 training/test cases for registration tasks to 1,251/209 for synthesis tasks. Segmentation tasks had median training and test set sizes of 175 and 72 cases, compared with 450/200 for classification and 420/153 for detection (Fig.~\ref{fig:fig3}a).

MRI (35\%) and CT (21\%) were the most frequently represented imaging modalities, followed by laparoscopy or endoscopy (13\%) and histopathology (11\%). Other modalities were only rarely represented (Fig.~\ref{fig:fig2}b). Compared with imaging utilization rates in the US \cite{smith2019trends}, MRI was 4.4-fold overrepresented, whereas ultrasound was severely underrepresented (0.16-fold; Suppl. Tab.~\ref{tab:suppl-tab3}). Median dataset size also varied across imaging modalities, ranging from 140/57 training/test cases for MRI to 1,905/177 for laparoscopy- or endoscopy-related tasks.

The imaging modality composition differed across problem categories (Fig.~\ref{fig:fig3}b). While MRI was the predominant modality for most problem categories, detection tasks were most frequently based on laparoscopy or endoscopy data (31\%). In contrast, classification tasks were distributed across several imaging modalities, with no single modality accounting for more than 21\% of tasks.

Tasks most frequently focused on the brain (30\%), followed by the eyes (10\%), colon (10\%), and lung (9\%), while many other anatomical regions accounted for only a small proportion of tasks (Fig.~\ref{fig:fig1}b).

Data were typically collected from a median of one center (mean: 6; maximum: 500) and a median of two devices (mean: 2.5, maximum: 18). Where reported, the median sex distribution was 52\% male and 47\% female (Suppl. Tab.~\ref{tab:suppl-tab6}). Only 26\% of tasks reported age information. Among those that did, the median reported mean age was 54 years (mean: 49.9, IQR: 13). Reported ages ranged from 0 to 100 years (Suppl. Tab.~\ref{tab:suppl-tab7}). 

\newpage

\newpage
\subsection{Accessibility and licensing practices limit responsible data reuse}
Although many datasets are nominally public, their accessibility and legal clarity often fall short of the expectations associated with the FAIR principles of Accessibility and Reusability. We found widespread barriers and inconsistencies in licensing and access conditions. 81\% of datasets were publicly available, but accessibility was often restricted by mandatory registration (38\%), organizer approval (20\%), or context limitations (9\%) such as restrictions on access or use by participants from certain countries or during specified periods.

Using the (Re)usable Data Project (RDP) framework \cite{carbon2019analysis}, we assessed the quality of data sharing on a 0-5 scale (5: license clearly permits unrestricted reuse and redistribution), reflecting the clarity of licensing terms, accessibility of the data, and restrictions on reuse. The median score across tasks was 3, corresponding to licenses that are clearly stated and of a standard type, yet include terms that substantially limit the (re)use or redistribution of the data. Notably, 46\% of tasks received a score of 2.5 or lower, reflecting difficulties in locating definitive licensing information, ambiguous terms that complicated interpretation, or access conditions requiring legal clarification (Suppl. Fig.~\ref{fig:suppl-fig1}.

Only 59\% of tasks provided sufficiently detailed license information and only 41\% used an unambiguous license. Creative commons (CC) licenses were applied in 51\%, although they were often modified or combined with a custom license (19\%). The most commonly used licenses were custom licenses (29\% of licenses), followed by CC BY (18\%), CC BY-NC-SA (18\%), CC BY-NC-ND (15\%), and CC BY-NC (11\%). Overall, restrictive licenses dominated (44\%), whereas permissive licenses remained comparatively rare (25\%). Only 40\% of peer-reviewed challenges applied the same license as approved by the conference chairs.

\begin{figure}[H]
    \centering
    \includegraphics[width=0.5\linewidth]{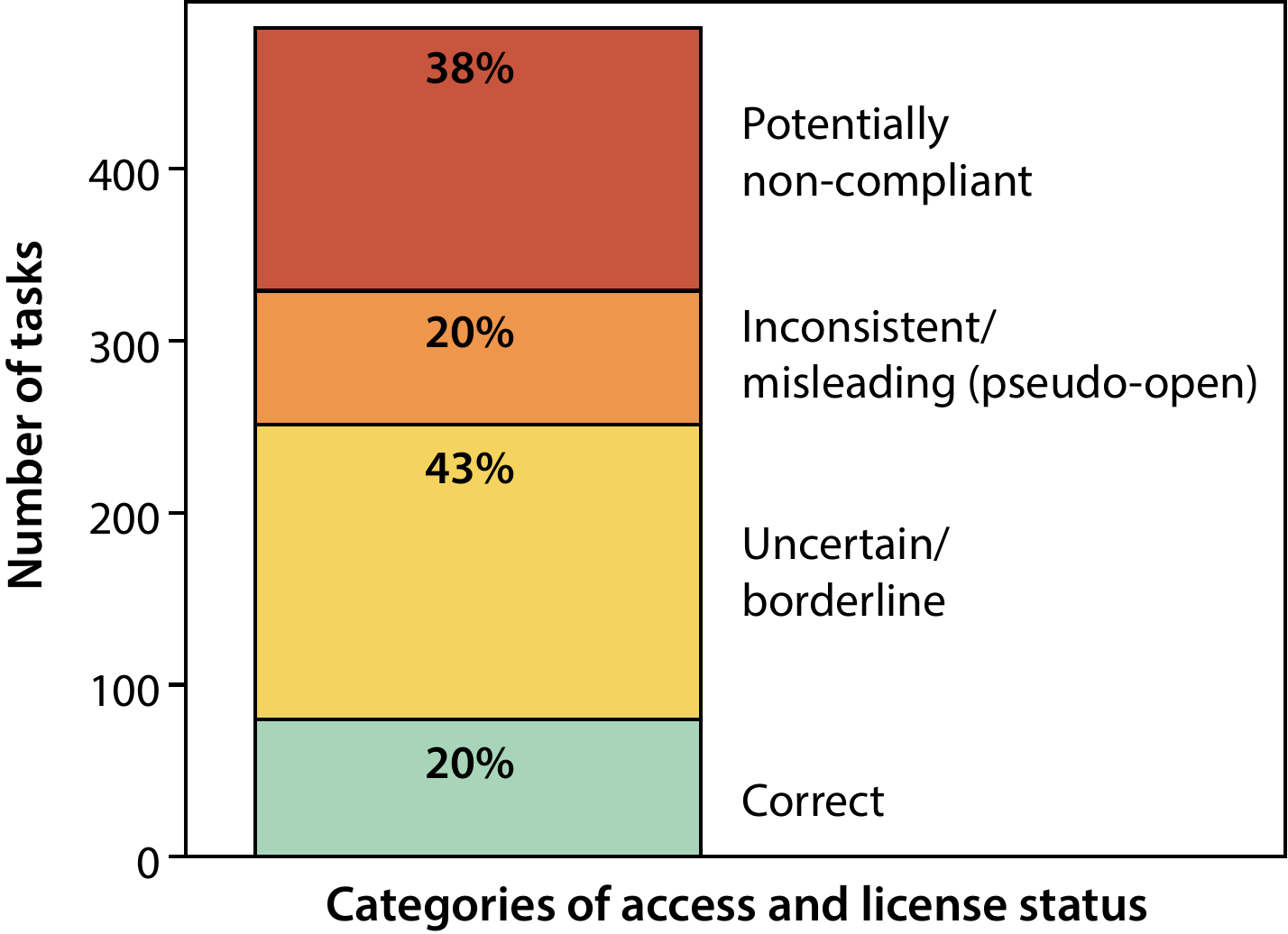}
    \caption{\textbf{Problematic practices in data licensing and access conditions undermine expectations associated with the FAIR principles of Accessibility and Reusability.}  Results are based on tasks with standard licenses or clearly interpretable licensing information (n = 398). Tasks were categorized according to observed licensing and access practices, ranging from correct implementations to unclear, misleading, and potentially non-compliant cases. Note that tasks may exhibit multiple types of issues, resulting in percentages exceeding 100\%.}
    \label{fig:fig4}
\end{figure}

\newpage
Based on tasks with standard or clearly interpretable licensing information, 80\% exhibited inconsistencies in data licensing or access practices, ranging from unclear or misleading conditions to clear cases of legal non-compliance, which we organized into three categories (Fig.~\ref{fig:fig4}, Tab.~\ref{tab:inconsistencies})\footnote{The classification is based on publicly available information provided by challenge organizers and associated publications. Due to differences in legal frameworks and the absence of case-specific legal review, the identified categories, particularly potentially non-compliant cases, reflect observed licensing practices and inconsistencies rather than definitive legal determinations.}. 

\begin{itemize}
    \item \textbf{Uncertain or borderline cases} (43\% of tasks) referred to situations in which data licensing or access conditions were formally permissible but lacked clarity or suitability (Tab.~\ref{tab:inconsistencies}; e.g., approval-based or context-limited access under restrictive licenses, use of NoDerivatives clauses, or the application of software-style licenses to datasets).
    \item \textbf{Inconsistent or misleading (pseudo-open) cases} (20\%) encompassed datasets that were presented as open but in practice imposing restrictions that contradicted the openness claim (Tab.~\ref{tab:inconsistencies}; e.g., approval or registration requirements despite open licenses, or declared licenses without accessible data).
    \item \textbf{Potentially non-compliant cases} (38\%) represented cases that suggest legal non-compliance (Tab.~\ref{tab:inconsistencies}; e.g., conflicting or contradictory license terms, conflicting license statements across sources, or explicit violations of license conditions).
\end{itemize}


\begin{footnotesize}
\captionof{table}{Overview of data access and license inconsistencies.}
\label{tab:inconsistencies}
\vspace{-1\baselineskip}
\begin{longtable}{ B{3.5cm} B{8.5cm} P{2.5cm} P{1cm}  }
 \toprule
 \multicolumn{4}{l}{\textbf{\textit{Correct}}} \\
\midrule
\textbf{Case} & \textbf{Description} & \textbf{Problem Type} & \textbf{Tasks} \\
\midrule
No obvious inconsistencies & The data license is clear, accessible, consistently applied across all sources, appropriate for the data type, and is consistent with stated reuse conditions. No ambiguities are present, including with respect to license scope, consistency, and stated reuse conditions. & None & 20\%  \\\midrule

\multicolumn{4}{l}{\textbf{\textit{Uncertain / borderline}}} \\
\multicolumn{4}{l}{\textit{Cases in which data licensing or access conditions are formally permissible but lack clarity, suitability, or}} \\
\multicolumn{4}{l}{\textit{transparency. These practices do not necessarily violate legal terms but complicate reuse and reproducibility.}} \\
\midrule
\textbf{Case} & \textbf{Description} & \textbf{Problem Type} & \textbf{Tasks} \\
\midrule

Platform registration under restrictive license & A dataset under a restrictive license requires registration before access. While this may be justified for ethical, legal, or governance reasons, it adds additional barriers and reduces transparency when no justification for the restriction is provided. It can be legally sound, but introduces a hurdle to usability and transparency. & Data access & 33\% \\ \hline

Additional data usage agreement mentioned & A dataset is accompanied by an additional data usage agreement that exists alongside the stated license. The presence of two parallel sets of terms introduces uncertainty about their relationship and whether the agreement merely clarifies or partially modifies the license. Such unclear interaction can, in some cases, result in overlapping or even contradictory obligations, creating potential legal risks for both data providers and users.  & 
Data access \& License & 23\%\\ \hline

Approval-based access under restrictive license & Approval is required for data under restrictive licensing conditions. Although this may align with the terms of the license, the absence of clear criteria creates uncertainty about permissible reuse. In addition, the absence of the researchers who managed the data (e.g., after defending theses) may make the dataset inaccessible. & Data access & 20\%\\\hline

NoDerivatives clause & A dataset is released under a license containing a NoDerivatives (ND) clause. In the context of AI and machine learning, it remains unclear whether model training (e.g., data augmentation) constitutes a derivative work, creating uncertainty about lawful use. & License & 18\%\\\hline

Context-limited access under restrictive license & Geographic, temporal, or event-related limitations are imposed under a restrictive license. These may be legitimate within a restricted-use framework, but they should be clearly specified to avoid confusion. The restriction is valid, but lack of clarity hinders reuse and may complicate interpretation of permissible use conditions. & Data access & 10\%\\\hline

Software-style license applied to dataset & A software license such as MIT or GNU General Public License (GPL) is applied to data rather than code. Because these licenses were not designed for datasets or databases, they fail to address relevant rights and create uncertainty about allowable reuse, which may be legally and practically critical. This choice of those licenses may not cover relevant rights (e.g., database rights), leading to an incomplete grant. & License & 1\%\\ \midrule

\multicolumn{4}{l}{\textbf{\textit{Inconsistent / misleading (pseudo-open)}}} \\
\multicolumn{4}{l}{\textit{Cases in which data sharing was presented as open but access or licensing conditions contradicted the}} \\
\multicolumn{4}{l}{\textit{openness claim. Such practices create an illusion of openness, undermining transparency and user trust.}} \\
\midrule
\textbf{Case} & \textbf{Description} & \textbf{Problem Type} & \textbf{Tasks} \\
\midrule

Platform registration despite open license & Access to a dataset that claims to be openly licensed still requires prior registration on a hosting platform. This contradicts the principle of unrestricted accessibility that defines open data and limits practical reuse. & Data access & 13\% \\ \hline

License provided but data unavailable & A license is publicly declared, but the corresponding dataset is no longer accessible, for example because a challenge has ended or a platform was shut down. Declaring a license without maintaining access misleads users and undermines reproducibility, as openness requires both permission and availability. & Data access \& License & 8\% \\ \hline

Approval-based access despite open license & Data usage requires explicit organizer approval even though the dataset is offered under an open license. Potential absence of the researchers who managed the data (e.g., after defending theses), may make the dataset even inaccessible. This turns open access into a permission-based process and potentially deters legitimate reuse. It is a misleading practice that undermines the core principle of an open license. & Data access & 5\% \\ \hline

Context-limited access despite open license & Access to an openly licensed dataset is limited to certain countries, time periods, or participation in a specific event. Such conditions contradict the universal reuse rights guaranteed by open licenses. & Data access & 3\% \\ \midrule

\newpage
\midrule
\multicolumn{4}{l}{\textbf{\textit{Potentially non-compliant}}} \\
\multicolumn{4}{l}{\textit{Cases involving licensing practices that raise concerns about legal or procedural compliance. They may}} \\
\multicolumn{4}{l}{\textit{conflict with commonly used licensing frameworks or stated license conditions and may introduce legal}} \\
\multicolumn{4}{l}{\textit{and ethical risks to data reuse.}} \\
\midrule
\textbf{Case} & \textbf{Description} & \textbf{Problem Type} & \textbf{Tasks} \\
\midrule

Conflicting license statements across sources & Different sources, such as the challenge or data website, challenge design document, and accompanying publication, provide conflicting licenses for the same dataset. This inconsistency leaves users uncertain which license terms apply and creates a risk of unintentional non-compliance due to reliance on an incorrect or outdated license. & License & 20\% \\ \hline

License missing or not findable & The dataset is made publicly available, but no explicit license or terms are provided on any source. In the absence of an explicit license, the default copyright status is "All Rights Reserved." Reuse beyond legally mandated exceptions may not be legally permitted. & License & 16\% \\ \hline

Conflicting or contradictory license terms & The same license contains mutually inconsistent conditions, such as a Creative Commons statement combined with additional restrictions. Such internal contradictions can make the licensing legally problematic and prevent users from determining which terms actually apply. & License & 3\% \\ \hline

Violation of general license terms & A data provider, organizer, or redistributor re-uses or distributes licensed data but fails to meet a core mandatory license requirement. Examples include not providing required attribution (BY) or re-licensing derivative works without the required Share-Alike (SA) clause. Such practice constitutes a breach of mandatory license terms and may result in copyright infringement. & License & 2\% \\ \hline

Incompatible license combination within a single source & A dataset is distributed within a single source under multiple license statements that impose mutually exclusive obligations, for instance both requiring attribution and prohibiting redistribution. These conditions may not be fulfilled simultaneously and can make lawful reuse difficult or impossible. & License & 2\% \\ \bottomrule

\end{longtable}
\end{footnotesize}

\subsection{Widespread reporting gaps hinder reusability and reproducibility}
\begin{figure}[h]
    \centering
    \includegraphics[width=0.8\linewidth]{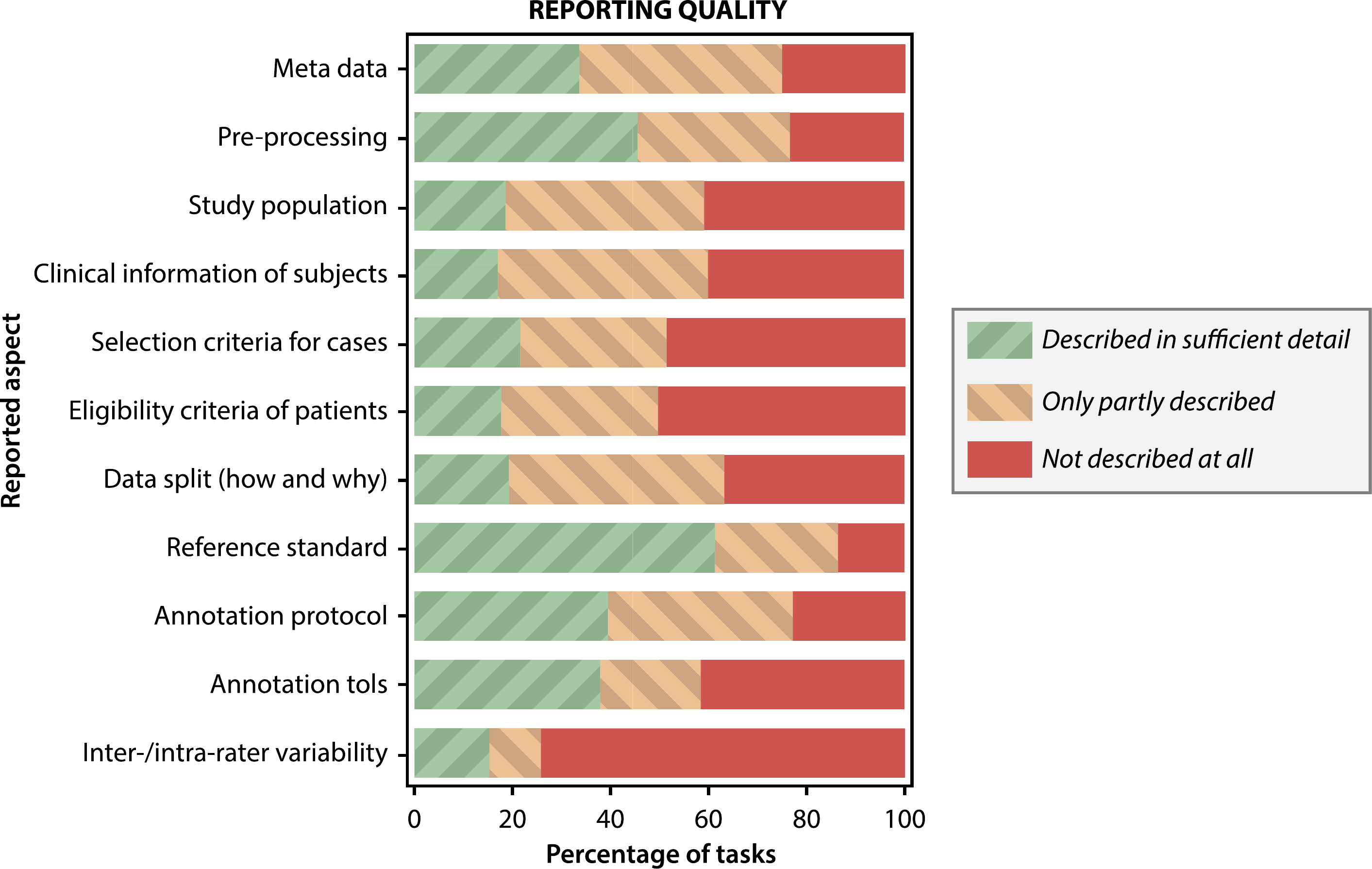}
    \caption{\textbf{Reporting quality related to data aspects in medical imaging AI competitions is poor.} Stacked barplots show the percentages of tasks in which key data aspects were described in sufficient detail (green), partly described (yellow), or not described at all (red).}
    \label{fig:fig5}
\end{figure}

Even where data access is granted, many datasets are not reusable in practice due to missing documentation (Fig.~\ref{fig:fig5}). Across multiple dimensions of reporting, we observed substantial gaps that undermine transparency and long-term usability.

\textbf{Technical and organizational information was frequently incomplete.} For 3\% of tasks, the challenge website was no longer functional. The type of award was unspecified in 43\% of tasks. Even when monetary awards were given, the exact prize amounts were missing in 56\% of cases. Information on the acquisition devices (e.g., scanner type) was absent in 43\% of tasks.

\textbf{Reporting on basic dataset organization and ethics approval was limited.} Only 19\% of tasks provided a sufficient description of the data splitting strategy, and metadata was not described at all in 25\% of tasks. De-identification procedures were not reported in 40\% of tasks, and 35\% of applicable tasks lacked information on ethics committee approval had been obtained.

\textbf{Information on cohort composition and study design was often missing.} The study population was not described in 41\% of tasks, and clinical information such as disease status, diagnosis, or treatment history per subjects was absent in 40\% of tasks. Case selection criteria were not reported in 49\% of tasks, and eligibility criteria were missing in 50\% of tasks. Although the overall country of data origin was often reported, the geographic composition of the training and test sets remained unspecified in 15\% and 13\% of tasks, respectively. Sex distribution was reported in only 30\% of tasks, and sufficiently detailed age distribution in only 5\%, with 26\% of tasks reporting partial information such as mean age. 

\textbf{Reporting related to data annotation was similarly sparse.} Annotation protocols were fully documented in only 40\% of tasks, annotation tools were not described in 42\%, and inter- or intra-rater variability was not reported in 74\%.

\section{Discussion}
Biomedical image analysis challenges are growing in number and visibility, shaping methodological development and performance standards. They aim to enable fair comparison and foster clinically translatable AI models. However, as their influence expands, so does their responsibility: benchmarks that fail to represent clinical diversity or that cannot be legally reused risk undermining fairness they aim to ensure.

Our analysis revealed that specific regions, modalities, and anatomical sites dominate challenges. This limited representation may constrain model generalizability and perpetuate gaps in geographic and clinical coverage. Notably, some patterns diverge markedly from clinical practice. For example, although MRI was the most common imaging modality in challenges (Fig.~\ref{fig:fig2}c), it accounts for a comparatively small share of clinical imaging. MRI imaging rates for older adults were only 139 per 1,000 person-years, compared to 428 for CT and 495 for ultrasound in the US in 2016 \cite{smith2019trends}. Restricting the comparison to these three modalities, MRI was 4.4-fold overrepresented in challenges relative to the reported clinical utilization distribution, whereas ultrasound was substantially underrepresented (ratio: 0.2). Beyond questions of clinical utility, this mismatch raises fairness concerns: prioritizing MRI-centric benchmarks implicitly centers resource-rich healthcare systems that can support high-cost modalities, thereby disadvantaging regions where ultrasound or X-ray are the primary diagnostic tools. This mismatch may reinforce the geographic underrepresentation of certain countries.

Similarly, segmentation was by far the most common task category across challenges, yet its direct clinical relevance is often constrained in real-world settings. Manual segmentation is rarely used in routine care, and segmented outputs primarily serve as intermediate steps such as for volume estimation or planning \cite{gremse2016imalytics}. Our additional analyses suggest that these task distributions are not solely driven by clinical priorities. Dataset sizes differed across problem categories, indicating that practical considerations related to data availability and annotation effort may influence which tasks are preferentially addressed in biomedical image analysis challenges. While our study design does not allow us to determine what drives these associations, several practical factors may plausibly contribute. Annotation requirements differ substantially between problem categories \cite{willemink2020preparing}. For example, some classification tasks require only image-level labels, whereas instance segmentation requires identifying and delineating individual objects. Other tasks may depend on data that are inherently more difficult to collect at scale, such as outcome prediction requiring longitudinal follow-up or linked clinical information. Dataset scale may also depend on examination frequency and the resources required for data acquisition and curation. For example, high-volume procedures such as cholecystectomy can generate large amounts of laparoscopic video data \cite{sundbom2014trends}, whereas assembling similarly sized datasets for rare diseases may require substantially longer acquisition periods. This interpretation is supported by the relationship between imaging modality and problem category. While MRI dominated most problem categories, detection tasks were considerably more common in laparoscopy and endoscopy than in other imaging modalities (Fig. 3b), suggesting that different imaging domains tend to emphasize different methodological questions and practical application scenarios. Regulatory, institutional, and political frameworks may further determine which data can be collected, accessed, and shared, shaping dataset availability for challenges. A similar interpretation may apply to anatomical regions. The predominance of brain imaging tasks likely reflects the historical availability of large, open-access neuroimaging datasets (e.g., \cite{varoquaux2022machine}) rather than population-level disease burden or unmet clinical needs. 

Overall, these patterns suggest that benchmarks may address problems that are readily measurable and feasible to study rather than those of greatest clinical relevance This challenges the implicit assumption that leaderboard success equates to clinical readiness \cite{gu2025illusion}. More generally, challenge leaderboards should always be interpreted with care. As discussed in prior work (e.g., \cite{maier2024metrics, maier2018rankings, wiesenfarth2021methods}), small differences in benchmark performance are often difficult to interpret and do not necessarily translate into clinically meaningful improvements. An excessive focus on leaderboard position may encourage optimization for potentially unsuitable benchmark-specific metrics rather than properties required for successful clinical deployment, such as robustness or workflow integration \cite{whitehill2018climbing}. More broadly, benchmark performance can serve as a valuable proxy for clinical translation when evaluated on representative and clinically meaningful benchmarks, as illustrated by the Digital Mammography DREAM Challenge. Clinical deployment, however, additionally depends on rigorous external validation, regulatory approval, and integration into clinical workflows.

We found that even when challenge data are nominally public, access is frequently restricted through organizer approval requirements, registration barriers, or ambiguous licensing. Registration or approval procedures may be justified for ethical, legal, or governance reasons, particularly for sensitive medical data. However, in combination with open licenses, additional access requirements should be avoided wherever possible, as they contradict the universal reuse rights such licenses guarantee. In extreme cases, datasets may become unreachable, for example after organizers leave their institutions, rendering datasets practically unavailable \cite{jimenez2024copycats}. Prior studies in clinical epidemiology have shown that in, "data available upon request" statements have extremely low compliance, similar to having no such statement at all \cite{gabelica2022many}. These issues are not administrative details; they actively prevent legitimate reuse and slow down scientific progress. Notably, \cite{christensen2019study} showed that articles that make their data available receive more citations in other fields such as economics and political science, underscoring the broader value of open data.

While the majority of tasks claim to provide public data, we found that more than 80\% include substantial reuse restrictions, especially in a sense of ambiguous licensing terms. Although additional data access restrictions can be legally valid, several challenges used impermissible licenses or terms. A common example includes conflicting license statements across different sources, such as stating a different license for the same dataset on the (potentially outdated) challenge website and the challenge publication. This creates uncertainty about which license governs reuse and may expose users to legal risks, including takedown requests, disputes, or claims arising from unintended violations. Other clear violations include conflicting or contradictory license terms, often induced by an additional data usage agreement, or the violation of general license terms, such as re-licensing works without a required Share-Alike (SA) clause. Beyond these clear violations, large grey zones persist. A common example is the usage of NoDerivatives (ND) clauses (e.g., CC-BY ND), where the notion of “derivative work” remains poorly defined in the context of AI \cite{peng2021mitigating}. For example, it can be unclear whether trained models, learned representations, data augmentation, or weights constitute derivative works under current copyright frameworks \cite{benjamin2019towards}. While studies in both AI law and EU copyright interpretation suggest that model training typically lacks the originality required for a new protected work unless the modification reflects creative input \cite{benjamin2019towards, de2018legal}, ND clauses may still formally prohibit reuse while offering no practical guidance for compliance.

Our results mirror a broader “crisis in misattribution” discussed in the general AI community \cite{longpre2024large}, where incomplete licensing information and inconsistent data provenance lead to widespread misattribution \cite{sourget2024citation} and uncertainty about permissible use. Limited data licensing expertise in the biomedical AI community may contribute to practices whose legal and practical implications are insufficiently considered \cite{sambasivan2021everyone}. For example, challenge organizers may introduce additional data usage agreements without recognizing potential conflicts with existing licenses or barriers for participants and future users. This is rarely due to negligence but rather limited access to specialized legal expertise or inconsistent institutional guidance, combined with ambiguities in interpreting license terms (e.g., commercial use or publication-related exemptions). These structural uncertainties become even more visible in practical conflicts: additional agreements may contradict open licenses, while attribution and Share-Alike requirements become nearly impossible to satisfy when models are trained on several datasets with incompatible terms \cite{szkalej2024generative}. For challenge organizers, these problems are amplified by restrictive Institutional Review Board (IRB) interpretations, cautious or inconsistent legal risk assessments that may diverge from evidence-based reuse risk evaluations, and insufficient data stewardship infrastructure, often leading to more restrictive or ad-hoc licensing than intended. At the same time, unclear or restrictive terms can substantially undermine reproducibility and long-term utility of benchmark datasets. That said, restrictive licenses or usage conditions can, in some situations, be necessary to preserve scientific integrity, and ensure ethical use. These access and licensing issues also fall short of FAIR expectations for clear terms, stable availability, and unambiguous conditions for responsible and practically feasible data reuse.

A related limitation is incomplete reporting in challenge publications, particularly of socio-demographic and clinical details. Basic descriptors such as age and sex were frequently missing, and clinical characterization was often incomplete. From a medical perspective, such information is essential for understanding the cohort and for assessing whether model performance is likely to generalize to real clinical populations. Similarly, insufficient reporting on study design, data splitting, acquisition devices, or annotation procedures restricts reproducibility and prevents meaningful comparison across challenges. Poor documentation therefore constitutes a fundamental obstacle to reusability, independent of access or licensing conditions.

The consequences for AI development are substantial: because biomedical AI depends heavily on benchmarks and challenge datasets that are frequently reused for subsequent method development, limited accessibility or unclear licensing conditions slow scientific progress and may distort how model performance is interpreted within and beyond leaderboard results. When benchmarks are unrepresentative or legally unusable, they lose their function as proxies for real-world performance and may inadvertently lead to misguided decisions, especially in safety-critical clinical settings. Ultimately, these problems raise the broader question of whether current benchmarking practices are aligned with real-world application needs or whether they increasingly serve academic signaling rather than supporting clinically meaningful AI development. Recent bans on studies using public health datasets such as NHANES \cite{o2025papers} illustrate that openness alone is not sufficient; openly available data can still be misused, which may appropriately trigger policy interventions, while highlighting how misuse and unclear governance can erode trust in open science ecosystems.

Our work comes with several limitations. The study was descriptive rather than prescriptive; due to substantial licensing gray zones and institution-specific rules, we did not aim to propose a universal guideline for challenge data governance. Our analysis was restricted to challenges conducted between 2018 and 2023, excluding recent editions that typically lack stable documentation or peer-reviewed publications. All information was extracted from publicly available sources, which were sometimes incomplete or outdated; missing information may therefore reflect either documentation gaps or limitations of what could be retrieved. Thus, we could not independently verify institutional or legal procedures (e.g., IRB approvals, internal data-sharing agreements), which limited our ability to assess compliance beyond what was reported. Although data extraction followed a structured double-blind process with 30 observers and conflict resolution, some assessments necessarily involved subjective interpretation. Our analysis focused on challenge datasets and documentation, without assessing downstream reuse, later publications, or the broader benchmark research lifecycle. Consequently, we cannot determine how observed dataset characteristics may influence subsequent methodological development or the trajectory of biomedical image analysis. Future work should investigate how benchmark datasets shape dataset reuse, research directions, and algorithm development. Meaningful external reference distributions were unavailable for several assessed dimensions. We therefore limited comparisons to geography and imaging modality, for which suitable reference data were available. However, establishing a reliable reference distribution for imaging modality remained challenging. Available studies on imaging utilization are often limited to specific countries or institutions and do not cover all modalities assessed in our study. Consequently, our modality comparison should be interpreted as an illustrative external proxy rather than a reference for global clinical practice. Finally, we did not assess the legal enforceability of identified licensing issues, which can vary across jurisdictions and may depend on context-specific factors beyond the scope of this study. Accordingly, our categorization reflects observed licensing practices and reported terms, rather than conclusions based on jurisdiction-specific legal standards or case-by-case legal interpretation.

Given the discussed problems, several emerging approaches aim to improve transparency, provenance, and legal certainty in dataset design. Complementary infrastructure-oriented work has proposed machine-readable, policy-as-code governance layers to decouple policy definition from enforcement, aiming to increase transparency, auditability, and legal clarity in decentralized AI data ecosystems \cite{kassem2025atechnical}. Related work from other areas highlighted that legal risks related to dataset reuse and redistribution cannot be assessed from license terms alone, but tracing dataset redistribution and lifecycle is essential \cite{kim2025not}. New technical efforts address this, including automated compliance agents and provenance-tracking systems such as the Data Provenance Explorer \cite{longpre2024large}. The Re(usable) Data Project offers a structured ruleset for assessing reusability, which we also applied in our analysis \cite{carbon2019analysis}. Licensing frameworks like the Montreal Data License provide AI-specific license language \cite{benjamin2019towards}. Complementary documentation frameworks such as "Datasheets for Datasets" \cite{gebru2021datasheets} outline standardized descriptions of dataset motivation, composition, and collection. Major data-hosting platforms have already begun to establish clearer licensing best practices that could serve as a model for biomedical challenges. For example, Hugging Face\footnote{\href{https://huggingface.co/docs/hub/en/repositories-licenses}{https://huggingface.co/docs/hub/en/repositories-licenses}} or Kaggle\footnote{\href{https://www.kaggle.com/discussions/general/116302 }{https://www.kaggle.com/discussions/general/116302}} provide overviews of acceptable license types, reuse conditions, and contributor responsibilities, although these aspects are not reviewed and users often leave respective documentation fields empty \cite{jimenez2024copycats}. Open collaborative efforts such as the GA4GH Data Use Ontology (DUO)\footnote{\href{https://www.ga4gh.org/product/data-use-ontology-duo/}{https://www.ga4gh.org/product/data-use-ontology-duo/}} and Croissant metadata standard\footnote{\href{https://research.google/blog/croissant-a-metadata-format-for-ml-ready-datasets/}{https://research.google/blog/croissant-a-metadata-format-for-ml-ready-datasets/}} further support this direction by providing controlled vocabularies and structured schemas for describing permissible data uses. Notably, schema.org\footnote{\href{https://research.google/blog/croissant-a-metadata-format-for-ml-ready-datasets/}{https://research.google/blog/croissant-a-metadata-format-for-ml-ready-datasets/}} already defines a standard license field, enabling consistent, machine-readable expressions of licensing terms. These initiatives address three identified issues: (1) difficulty identifying license types, (2) confusion between access restrictions and use permissions, and (3) inconsistent interpretation of use cases, including geographic limitations. Integrating these standards into biomedical image analysis challenges could substantially reduce ambiguity and improve data governance transparency.

One step toward improving the usability and fairness of challenge datasets would be the introduction of baseline requirements for metadata and licensing. Leading conferences such as MICCAI and ISBI could require challenge organizers to provide machine-readable metadata and clear, standardized licensing information, for example through the RDP or Croissant schemas. Establishing such requirements during submission and peer review would reduce ambiguity, support lawful reuse, and strengthen the long-term value of challenge datasets.

Beyond technical standards for metadata and licensing, improving future challenges also requires more comprehensive and targeted reporting. Motivated by the findings of this work, we are currently in the process of updating the BIAS (Transparent Reporting of Biomedical Image Analysis Challenges) reporting guideline \cite{maier2020bias} through an international Delphi consensus process. The revision extends the original checklist with additional items identified in this study, including data accessibility, justification for licensing, dataset representativeness, demographic and clinical characteristics, potential confounders, and fairness-related considerations. We hope that integrating these aspects into an established reporting standard will facilitate their broad adoption by future challenge organizers.

Overall, our findings indicate that current challenges fall far short of these emerging standards and therefore do not yet provide legally or clinically meaningful benchmarks for real-world translation. Our intention is not to criticize individual challenge organizers or prescribe how biomedical image analysis challenges should be organized. Many of the limitations identified in this work reflect broader structural constraints, including limited resources, restricted access to diverse clinical data, and legal or institutional barriers to data sharing that are largely beyond the control of individual organizers. Instead, we aim to raise awareness of these limitations, provide an empirical basis for discussion, and encourage feasible improvements in reporting quality, metadata documentation, and licensing practices. We hope these findings contribute to more transparent, reusable, and clinically meaningful benchmarking practices.

\section{Methods}
In this study, we use the terms “challenge” and “task” as defined by \cite{maier2018rankings}. A "challenge" denotes an open competition on a dedicated scientific problem in biomedical image analysis, typically organized by a consortium issuing a call for participation and potentially comprising multiple tasks. A “task” refers to a subproblem within a challenge for which a separate assessment or leaderboard is provided, potentially with its own performance metrics.

\subsection{Inclusion criteria}
Our aim was to capture all biomedical image analysis challenges that have been conducted between 2018 and 2023. Earlier years were not included because challenges up to 2016 had already been systematically reviewed in prior work \cite{maier2018rankings}, and our study focused on more recent developments in challenge design. We did not include 2024 or 2025 challenges either, as our focus was on information provided in scientific papers, which may have a delay of more than a year to be published after challenge execution.

To acquire the data, we analyzed the websites hosting/representing biomedical image analysis challenges, namely grand-challenge.org, dreamchallenges.org, and kaggle.com as well as websites of main conferences in the field of biomedical image analysis, namely Medical Image Computing and Computer Assisted Intervention (MICCAI), International Symposium on Biomedical Imaging (ISBI), Conference on Neural Information Processing Systems (NeurIPS), International Society for Optics and Photonics (SPIE) Medical Imaging, International Conference on Image Analysis and Recognition (ICIAR), Medical Imaging with Deep Learning (MIDL), and the annual meeting of the Radiological Society of North America (RSNA). Three challenges were excluded as their focus was not biomedical, or because no publicly accessible website or publication could be identified. This yielded a list of 241 challenges comprising 458 tasks.

\subsection{Systematic screening}
A list of parameters for the screening was created in several rounds of iterations. Parameters included items related to general information (such as challenge website, associated venues, or number of participants), information on data access and licenses (including concrete licenses and bad practices), and information on data (quality), including dataset sizes, geographical regions origin of data, or reporting quality of key data-related aspects.

Two independent observers reviewed each challenge for these parameters, with a total of thirty screeners. Afterwards, a third senior screener compared results and resolved conflicts to a final decision. Conflicts were resolved according to predefined guidelines established within the conflict-resolution team, ensuring consistent interpretation of incomplete or contradictory information across screeners. In total, three senior screeners participated in this adjudication step.

\subsection{Categorization of data access and licensing practices}
In addition to the extraction of predefined screening parameters, categories to describe recurring data access and licensing practices observed across challenges were identified. The categorization was developed iteratively based on patterns identified during screening and was intended to structure and describe common practices rather than to establish a legal standard or assess legal enforceability. 

Category definitions and descriptions were refined through internal discussion among co-authors to ensure clarity, internal consistency, and applicability across a wide range of challenges. The final categorization was reviewed by a legal expert to identify ambiguous or potentially misleading formulations and to ensure that descriptions accurately reflected observed practices without making jurisdiction-specific legal determinations.

\subsection{Quantification of over- and underrepresentation}
To quantify over- and underrepresentation, we compared observed challenge data with suitable external reference points. Representation ratios were calculated as the proportion observed in challenge datasets divided by the corresponding proportion in the empirical reference distribution. Values greater than 1 indicate overrepresentation, whereas values below 1 indicate underrepresentation.

For geographical representation, the reference distribution was based on the global population distribution by continent in 2020 obtained from \cite{sadigov2022rapid}. For imaging modalities, no single published reference providing comparable utilization rates across all imaging modalities identified in our study was available. Rather than combining estimates from different sources, we used the published clinical imaging utilization rates for MRI, CT, and ultrasound reported by \cite{smith2019trends} as a consistent reference. Proportions were normalized to these three modalities before calculating representation ratios.

Throughout the manuscript, we use the terms “over- and underrepresentation” only with respect to an explicitly specified reference point or distribution, while other differences in dataset composition are reported descriptively.

\section*{Acknowledgements}
This work was initiated by the Helmholtz Association of German Research Centers in the scope of the Helmholtz Imaging Incubator (HI), the MICCAI Special Interest Group on biomedical image analysis challenges, the benchmarking and evaluation working group of the MONAI initiative, and the Grand Challenge team. 

We would like to thank Omar Benjelloun (Google DeepMind, Paris, France), Enzo Ferrante (Instituto de ciencias de la
computación, CONICET - Universidad de Buenos Aires, Argentina), Nitisha Jain (King’s College London, London, UK), Anne Mickan (Department of Medical Imaging, Radboud University Medical Center, Nijmegen, The Netherlands), Marco Nolden (Division of Medical Image Computing , German Cancer Research Center (DKFZ), Heidelberg, Germany; Helmholtz Metadata Collaboration (HMC)), Özlem Özkan (Helmholtz Metadata Collaboration (HMC)) for their additional input and feedback on the manuscript. 

We would like to thank Thomas Li, Michael Kim, Yihao Liu, Karthik Ramadass, Gaurav Rudravaram, Shunxing Bao, Adam Saunders, Lianrui Zuo, Lucas Remedios, Chenyu Gao, Andre T.S. Hucke, Trent Schwartz, Mike Phillips from the Vanderbilt University, Nashville, Tennessee, USA, for their help in screening challenges.

\section*{Funding Statement}
This work was supported by the Helmholtz Association of German Research Centers in the scope of the Helmholtz Imaging Incubator (HI). The work was further supported by the German Federal Ministry of Education and Research (DECIPHER-M, 01KD2420F). Research reported in this publication was partially supported by the Informatics Technology for Cancer Research (ITCR) funding program of the National Cancer Institute (NCI) of the National Institutes of Health (NIH) under award numbers U01CA242871 and U24CA279629 (S.B.). The content of this publication is solely the responsibility of the authors and does not represent the official views of the NIH or any other funding body.  V.C. received support from Novo Nordisk Foundation NNF24OC0092612. O.C., M.S.: The research leading to these results has received funding from the French government under management of Agence Nationale de la Recherche as part of the “France 2030” program (reference ANR-23-IACL-0008, project PRAIRIE-PSAI), as part of the "Investissements d'avenir" program (reference ANR-10-IAIHU-06, project Agence Nationale de la Recherche-10-IA Institut Hospitalo-Universitaire-6), and from the European Union’s Horizon Europe Framework Programme (grant number 101136607, project CLARA). K.F. was funded by the PPP Allowance made available by Health Holland, Top Sector Life Sciences \& Health. M.K. and M.M.: This research was supported by the Czech Ministry of Education, Youth and Sports (Projects LM2023050 and CZ.02.01.01/00/23\_015/0008205). T. R. was supported by a scholarship from the Hanns Seidel Foundation with funds from the Federal Ministry of Education and Research Germany (BMBF). A.J.-S. has received funding from the Independent Research Council Denmark (DFF) Inge Lehmann 1134-00017B. D.S.: This research has been funded by the Dutch Research Council. 

\section*{Author Contributions Statement}
A.R. initiated and led the study, wrote and reviewed the manuscript, designed the parameter list for screening, categorized licenses, screened challenges, served as a conflict resolver for the challenge screening, designed the experiments, analyzed the data, and designed all figures. 
E.C. screened challenges, served as a conflict resolver for the challenge screening, designed the experiments, wrote and reviewed the manuscript.
S.S. screened challenges, co-designed the parameter list for screening, and reviewed the manuscript.
A.E.K. screened challenges, served as a conflict resolver for the challenge screening, designed the experiments, and reviewed the manuscript.
K.F. screened challenges, co-designed the parameter list for screening, and reviewed the manuscript.
D.H. screened challenges, co-designed the parameter list for screening, and reviewed the manuscript.
B.A.L. screened challenges and reviewed the manuscript.
C.S. screened challenges and reviewed the manuscript.
N.He. screened challenges and reviewed the manuscript.
S.L. screened challenges and reviewed the manuscript.
M.M. screened challenges and reviewed the manuscript.
M.S. screened challenges and reviewed the manuscript.
A.Y.-P. screened challenges and reviewed the manuscript.
V.B. screened challenges and reviewed the manuscript.
Ö.S. screened challenges and reviewed the manuscript.
S.d.J. screened challenges and reviewed the manuscript.
M.K. screened challenges and reviewed the manuscript.
T.R. screened challenges and reviewed the manuscript.
N.Ha. screened challenges and reviewed the manuscript.
F.M.-G. categorized licenses, provided legal advice, and reviewed the manuscript.
S.H. screened challenges and reviewed the manuscript.
M.A.R. screened challenges and reviewed the manuscript.
A.Sa. screened challenges, co-designed the parameter list for screening, and reviewed the manuscript.
V.T. screened challenges and reviewed the manuscript.
P.H. screened challenges and reviewed the manuscript.
A.J.-S. screened challenges, co-designed the parameter list for screening, and reviewed the manuscript.
Q.Y. screened challenges and reviewed the manuscript.
V.C. screened challenge, co-designed the parameter list for screening, and reviewed the manuscript.
S.B. screened challenges and reviewed the manuscript.
A.Se. categorized licenses and reviewed the manuscript.
S.B. screened challenges and reviewed the manuscript.
A.K. screened challenges and reviewed the manuscript.
K.V.V. initiated the study, co-designed the parameter list for screening, and reviewed the manuscript.
B.v.G. initiated the study and reviewed the manuscript.
L.M.-H. initiated and led the study and wrote and reviewed the manuscript.

\section*{Competing Interests Statement}
O.C. reports having received consulting fees from Therapanacea. O.C. reports that other principal investigators affiliated to the team which he co-leads have received grants (paid to the institution) from Sanofi and Biogen. OC reports that his spouse was an employee of myBrainTechnologies and is an employee of DiamPark. B.v.G. has stock in Thirona and Plain Medical. A.S. received lecture honoraria for AI-assisted prostate cancer diagnosis from Guerbet, and holds stocks in NVIDIA Corporation. We acknowledge Mr. H. J. M. Roels’s financial support through a donation to Oncode Institute. The remaining authors declare no competing interests.

\newpage
\bibliography{references}
\newpage
\section*{Supplementary Information}
\setcounter{figure}{0}
\setcounter{table}{0}
\renewcommand{\figurename}{\textbf{Suppl. Figure}}
\renewcommand{\tablename}{\textbf{Suppl. Table}}

\renewcommand{\thefigure}{\textbf{\arabic{figure}}}
\renewcommand{\thetable}{\textbf{\arabic{table}}}

\renewcommand*{\thesection}{Supplementary Note \arabic{section}:}
\renewcommand*{\thesubsection}{\arabic{section}.\arabic{subsection}:}

\appendix
\section{Number of challenge participants}
\vspace{-1cm}
\begin{table}[H]
\centering
\caption{Descriptive statistics on number of (final) participants across challenge venues (if reported) with at least two tasks being organized at the venue within the screening period. Abbreviations: Interquartile Range (IQR); Medical Image Computing and Computer Assisted Intervention (MICCAI); International Symposium on Biomedical Imaging (ISBI); Conference on Neural Information Processing Systems (NeurIPS); Medical Imaging with Deep Learning (MIDL); International Society for Optics and Photonics (SPIE) Medical Imaging; International Conference on Image Analysis and Recognition (ICIAR).}
\label{tab:suppl-tab1}
\begin{tabularx}{0.8\textwidth}{l *{5}{>{\centering\arraybackslash}X}}
\toprule
\textbf{Venue} & \textbf{Minimum} & \textbf{Mean} & \textbf{Median} & \textbf{IQR} & \textbf{Maximum} \\
\midrule
MICCAI  & 0  & 45.2  & 11   & 13  & 3{,}308 \\
ISBI    & 2  & 18.6  & 14   & 14  & 62 \\
Kaggle  & 14 & 1{,}289 & 1{,}347 & 962 & 2{,}078 \\
NeurIPS & 3  & 15.5  & 19   & 10  & 28 \\
MIDL    & 6  & 18    & 18   & 12  & 30 \\
SPIE    & 39 & 55    & 55   & 16  & 71 \\
ICIAR   & 4  & 25.3  & 8    & 45  & 64 \\
\bottomrule
\end{tabularx}
\end{table}

\section{Quantification of over- and underrepresentation}
\vspace{-1cm}
\begin{table}[H]
\centering
\caption{Quantification of geographical over- and underrepresentation of challenge data. The global population distribution as of 2020 \cite{sadigov2022rapid} was used as an empirical proxy for geographic representation. The representation ratio was calculated as the proportion observed in challenge datasets divided by the corresponding proportion in the empirical reference distribution. Values greater than 1 indicate overrepresentation, whereas values below 1 indicate underrepresentation. }
\label{tab:suppl-tab2}
\begin{tabularx}{0.8\textwidth}{l *{4}{>{\centering\arraybackslash}X}}
\toprule
\textbf{Continent} & \textbf{Global population} & \textbf{Challenge distributon} & \textbf{Ratio} \\
\midrule
Asia & 59.6\% & 28.4\% & 0.48 \\
Africa & 17.2\% & 1.0\% &0.06 \\
Europe & 9.6\% & 75.8\% & 7.91 \\
South America & 8.4\% & 1.3\% & 0.16\\
North America & 4.7\% & 75.3\% &15.9 \\
Oceania & 0.6\% & 2.4\% & 4.35\\
\bottomrule
\end{tabularx}
\end{table}

\newpage
\begin{table}[H]
\centering
\caption{Quantification of modality-related over- and underrepresentation of challenge data. Published imaging utilization rates for MRI, CT, and ultrasound in the US as of 2016 \cite{smith2019trends} were used as an empirical proxy for modality representation. The representation ratio was calculated as the proportion observed in challenge datasets divided by the corresponding proportion in the empirical reference distribution. Proportions were calculated among MRI, CT, and ultrasound only. Values greater than 1 indicate overrepresentation, whereas values below 1 indicate underrepresentation.}
\label{tab:suppl-tab3}
\begin{tabularx}{0.8\textwidth}{l *{4}{>{\centering\arraybackslash}X}}
\toprule
\textbf{Modality} & \textbf{Clinical imaging utilization} & \textbf{Challenge distributon} & \textbf{Ratio} \\
\midrule
Ultrasound & 46.6\% & 7.3\% & 0.16\\
CT & 40.3\% & 34.7\% & 0.86 \\
MRI & 13.1\% & 58.0\% & 4.43 \\
\bottomrule
\end{tabularx}
\end{table}

\section{Challenge dataset sizes}
\vspace{-1cm}
\begin{table}[H]
\centering
\caption{ Descriptive statistics on data sample sizes across data subsets, if reported. Abbreviation: Interquartile Range (IQR)}
\label{tab:suppl-tab4}
\begin{tabularx}{0.65\textwidth}{l *{3}{>{\centering\arraybackslash}X}}
\toprule
 & \textbf{Training} & \textbf{Validation} & \textbf{Test} \\
\midrule
\textbf{Reported by (\%)} & 93.9\% & 91.2\% & 86.2\% \\
\textbf{Minimum}    & 0       & 1              & 1 \\
\textbf{Mean}   & 4{,}246.5 & 411.9     & 704.8 \\
\textbf{Median} & 288     & 80             & 102 \\
\textbf{IQR}    & 1{,}130 & 170            & 324 \\
\textbf{Maximum}    & 530{,}706 & 18{,}368  & 47{,}227 \\
\bottomrule
\end{tabularx}
\end{table}

\newpage
\vspace{-5cm}
\begin{table}[H]
\centering
\caption{Median and Interquartile Range (IQR) of training and test set sizes by country, ordered by the frequency with which each country contributed data to the included challenges. IQR values are shown in brackets. Only countries represented in at least one challenge are shown. Median and IQR values were calculated using only tasks for which the country-specific training and test split was explicitly reported. NA indicates that no such information was available for the respective country.}
\label{tab:suppl-tab5}
\begin{tabularx}{0.8\textwidth}{l *{2}{>{\centering\arraybackslash}X}}
\toprule
\textbf{Country} & \textbf{Training} & \textbf{Test} \\
\midrule
US & 210.0 (25.0; 452.5) & 100.0 (28.3; 255.5) \\
China & 400.0 (127.5; 743.8) & 150.0 (70.0; 400.0) \\
Germany & 109.0 (31.0; 2000.0) & 40.0 (21.3; 200.0) \\
The Netherlands & 113.0 (34.0; 360.0) & 85.5 (27.0; 140.8) \\
UK & 147.5 (17.5; 600.0) & 52.0 (15.0; 350.0) \\
Switzerland & 85.0 (73.3;  90.0) & 50.0 (35.0; 53.0) \\
France & 90.0 (40.0; 92.5) & 60.0 (42.0; 60.0) \\
Italy & 22,601.0 (11,474.0; 22,601.0) & 6,223.0 (6,223.0; 6,223.0) \\
India & 413.0 (298.0; 413.0) & 190.0 (103.0; 385.8) \\
Canada & 201.0 (201.0; 201.0) & 98.0 (27.0; 422.0) \\
Hungary & NA & NA \\
Spain & 78.0 (35.3; 136.5) & 100.0 (100.0; 115.0) \\
Norway & 22.5 (22.0; 23.0) & 10.0 (10.0; 10.0) \\
Australia & 77.0 (77.0; 77.0) & 25.0 (25.0; 25.0) \\
Turkey & 20.0 (20.0; 40.0) & 20.0 (20.0; 40.0) \\
Portugal & 235.0 (235.0; 297.0) & 58.0 (58.0; 79.0) \\
Sweden & 750.0 (750.0; 917.5) & 300.0 (250.0; 315.0) \\
Israel & 60.0 (60.0; 60.0) & 14.0 (13.5; 14.5) \\
Belgium & NA & NA \\
South Korea & 50.0 (47.8; 50.0) & 40.0 (25.0; 40.0) \\
Austria & 69.5 (54.75; 84.25) & 40.0 (40.0; 40.0) \\
Denmark & 1,466.5 (409.0; 2,524.0) & 20.0 (20.0; 20.0) \\
Brazil & NA & NA \\
Czech Republic & NA & NA \\
Russia & NA & NA \\
Kazakhstan & 1,000.0 (1,000.0; 1,000.0) & 300.0 (300.0; 300.0)\\
Taiwan & 288.0 (288.0; 288.0) &  180.0 (180.0; 180.0) \\
Poland & NA & NA \\
New Zealand & 40.0 (40.0; 40.0) & 20.0 (20.0; 20.0) \\
Slovenia & 42.0 (42.0; 42.0) & 14.0 (14.0; 14.0) \\
Egypt & NA & NA \\
Chile& NA & NA \\
Africa & 120.0 (120.0; 120.0) & 60.0 (60.0; 60.0) \\
Singapore & 286.0 (286.0; 286.0) & 128.0 (128.0; 128.0) \\
Nigeria & NA & NA \\
Ireland & NA & NA \\
Japan & 17.0 (17.0; 17.0) & 10.0 (10.0; 10.0) \\
Morocco & NA & NA \\
Malta & NA & NA \\
Bosnia & NA & NA \\
Thailand & NA & NA \\
Argentina & NA & NA \\
\bottomrule
\end{tabularx}
\end{table}

\section{Data-sharing scores}
\begin{figure}[H]
    \centering
    \includegraphics[width=0.6\linewidth]{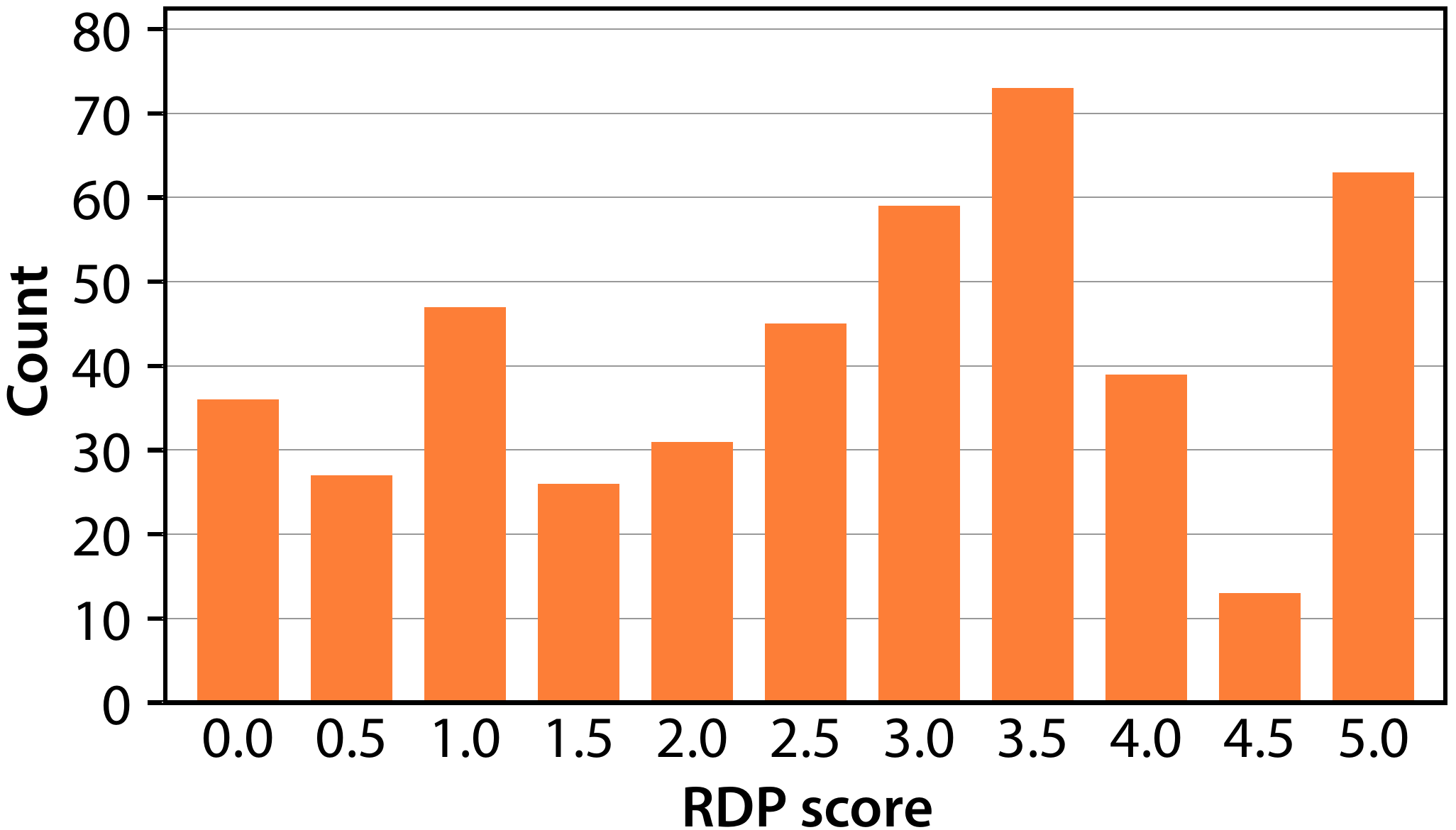}
    \caption{Distribution of data-sharing scores across tasks (n = 458 tasks). Data-sharing quality was assessed on a 0–5 scale following the (Re)usable Data Project (RDP) framework \cite{carbon2019analysis}, where higher scores indicate clearer licensing terms, greater data accessibility, and fewer restrictions on reuse and redistribution. Bars show the number of tasks assigned to each score.}
    \label{fig:suppl-fig1}
\end{figure}

\section{Challenge data demographics}
\vspace{-1cm}
\begin{table}[H]
\centering
\caption{Descriptive statistics on sex distribution across tasks, if reported. The proportion of tasks in which sex was reported is shown, together with descriptive statistics per sex. Abbreviation: Interquartile Range (IQR)}
\label{tab:suppl-tab6}
\begin{tabularx}{0.5\textwidth}{l *{2}{>{\centering\arraybackslash}X}}
\toprule
 & \textbf{Male} & \textbf{Female} \\
\midrule
\textbf{Reported by (\%)} & 29.9\% & 29.9\% \\
\textbf{Min}    & 0.0\%  & 0.0\% \\
\textbf{Mean}   & 47.6\% & 52.4\% \\
\textbf{Median} & 52.0\% & 47.0\% \\
\textbf{IQR}    & 26.0\% & 26.0\% \\
\textbf{Max}    & 100.0\% & 100.0\% \\
\bottomrule
\end{tabularx}
\end{table}

\newpage
\begin{table}[H]
\centering
\caption{Descriptive statistics of reported age information across tasks. For each task, summary age descriptors (minimum age, mean age, median age, and maximum age) were extracted when available. For each descriptor, the proportion of tasks in which it was reported and descriptive statistics computed across the reported values only are shown. Abbreviation: Interquartile Range (IQR)}
\label{tab:suppl-tab7}
\begin{tabularx}{0.85\textwidth}{l *{4}{>{\centering\arraybackslash}X}}
\toprule
 & \textbf{Minimum age} & \textbf{Mean age} & \textbf{Median age} & \textbf{Maximum age} \\
\midrule
\textbf{Reported by (\%)} & 19.0\% & 18.3\% & 7.0\% & 18.1\% \\
\textbf{Minimum}    & 0.0  & 0.0  & 0.0  & 0.0 \\
\textbf{Mean}   & 22.0 & 49.9 & 50.5 & 73.3 \\
\textbf{Median} & 19.0 & 54.0 & 60.0 & 83.0 \\
\textbf{IQR}    & 8.0  & 13.0 & 16.0 & 23.0 \\
\textbf{Maximum}    & 55.0 & 72.0 & 66.0 & 100.0 \\
\bottomrule
\end{tabularx}
\end{table}

\end{document}